\documentclass[11pt]{article}

\usepackage{acl}

\usepackage{amsmath}
\usepackage{times}
\usepackage{latexsym}
\usepackage{booktabs}
\usepackage[normalem]{ulem}
\usepackage{float}
\usepackage{caption}

\usepackage[T1]{fontenc}
\usepackage{tabularx}
\usepackage{colortbl}
\usepackage[utf8]{inputenc}
\usepackage{microtype}

\usepackage{inconsolata}

\usepackage{graphicx}
\usepackage{todonotes}
\title{Which Forms of Caregiver Feedback Support Grammar Learning? \\A Reinforcement-Learning Study of Child-Like Language Models}

\author{
  \textbf{Jing Liu\thanks{Equal contribution.}}
  \qquad
  \textbf{Marianne Schweitzer\footnotemark[1]}
  \qquad
  \textbf{Abdellah Fourtassi}
  \\[0.5em]
  ENS, Université PSL, EHESS, CNRS, Paris, France
  \\
  \texttt{jing.liu@psl.eu}
  \\[0.5em]
  Aix Marseille Université, CNRS, LIS, Marseille, France
  \\
  \texttt{marianne.schweitzer@lis-lab.fr}
  \qquad
  \texttt{abdellah.fourtassi@lis-lab.fr}
}

\begin{document}
\maketitle

\begin{abstract}

Social interaction is central to children’s language learning, but the  effects of different forms of caregiver feedback are difficult to isolate in naturalistic data.
We use child-like language models as controlled learners to test which forms of feedback support grammatical development. Small GPT-2-style models are pretrained on child-directed language from CHILDES, then fine-tuned with reinforcement learning using reward models trained to capture four feedback types: communicative feedback, structural alignment, semantic contingency, and affective feedback. Reward fine-tuning yields limited gains on minimal-pair evaluations, but clearer effects in free generation. Structural alignment produces the strongest improvements in grammaticality, providing a novel, plausible mechanistic account of how this feedback can support grammar learning. Communicative feedback yields more moderate gains. In contrast, semantic contingency and affective feedback do not improve grammaticality, although further analyses suggest that they may support other aspects of language learning beyond grammar. These results suggest that different forms of caregiver feedback make complementary contributions to language learning.



\end{abstract}

\section{Introduction}

Children do not learn language from exposure to input alone, but also through social interaction in which caregivers respond to their produced utterances \cite{fusaroli_caregiver_2023, masek_where_2021, clark_conversational_2020}. A long-standing question in language acquisition concerns the role of such feedback in shaping grammatical development, especially if caregiver responses can  help the learner know when their productions were well- or ill-formed \cite[see review in][]{, sokolov_changing_1994}. 

Although caregivers rarely provide explicit correction of grammatical errors \cite{brown_derivational_1970}, researchers have proposed that they nevertheless provide global, noisy feedback from which children can infer whether their utterances were likely grammatical or ungrammatical \cite[e.g.,][]{hirsh-pasek_brown_1984, demetras_feedback_1986, sokolov_changing_1994}. In this view, the child can be understood as learning in a reinforcement-like fashion, using the valence of the feedback as positive or negative rewards that guide which productions to maintain or adjust in future interactions \cite[e.g., see][]{whitehurst_what_1988, nikolaus_communicative_2023, schoneberger2010three}.

The developmental literature has documented several forms of caregiver response that can be interpreted, within a reinforcement-learning framework, as valenced feedback. The most direct examples are \textit{communicative feedback} cues, namely a) acknowledgments (e.g., "okay", "yeah"), which signal communicative success and thus increase the likelihood that the child's sentence was well-formed, and b) clarification requests (e.g., "huh?", "what?"), which signal communicative failure or misunderstanding and thus increase the likelihood that the sentence was ill-formed \cite{clark_conversational_2020, demetras_feedback_1986, newport_mother_1977, saxton_prompt_2005, nikolaus_communicative_2023a, nikolaus_communicative_2022}. 

A second relevant class is \textit{semantic contingency}, or the extent to which caregiver responses remain responsive to the child's topic and intended meaning  \cite{hoff-ginsberg_topic_1987, che2018assessing, hirsh2015contribution, agrawal_automatic_2024, agrawal2026scaffolding}. The degree of this contingency captures the feedback valence: low contingency in the caregiver response signals semantic misalignment and thus suggests that the child’s utterance may be ill-formed.

There are other forms of caregiver feedback that also provide such valenced signals, even if they have not traditionally been associated with this learning framework. Perhaps most directly relevant to grammar learning is \textit{structural alignment} \cite{fusaroli_caregiver_2023, dale_unraveling_2006, fernandez_quantifying_2014, misiek_development_2020}, whereby caregivers respond by reusing the child's syntactic frame while potentially introducing different lexical content (e.g., child: “I eat” (Pronoun-Verb) → caregiver: “Mommy eats too”). Here, the feedback valence is reflected in whether such alignment occurs, signaling support for the child’s use of that syntactic frame.

A final type of feedback is \textit{affective feedback}, such as whether caregivers respond with positive affect (e.g., "that's amazing!") versus a more emotionally neutral response \cite{rivero2023relations}.
 
These cues are far more available to children than explicit grammatical corrections, making their role in language learning much more plausible \cite{demetras_feedback_1986, hirsh-pasek_brown_1984}. Yet, they provide only indirect and noisy learning signals. First, they are not always directed at grammar specifically. Second, even in cases when grammar is the main target, the signal is global (i.e., utterance-level) and does not identify the specific error within the utterance. This makes \textit{credit assignment} a central challenge \cite{marcus_negative_1993, pinker_learnability_1989, saxton_contrast_1997}, and also makes interactive learning from caregiver feedback an interesting \emph{empirical} case.

Nevertheless, little is known about how these different feedback types contribute to children’s grammatical learning. 
While corpus studies allow a quantitative account of caregivers' feedback behavior, \textit{the causal effect} of each type of feedback or reward on learning remains difficult to establish. A common methodological issue facing previous attempts \cite[e.g., see discussion in][]{saxton_negative_2005} is that different types of feedback are provided to the same child, making it difficult---and ethically problematic---to isolate their effect over the developmental timescales on which language is learned.


Recent computational work has begun to revisit these questions by using LMs as controlled learning environments for testing hypotheses about language learning, thereby combining the targeted precision of experimental methods with the ecological validity of corpus studies \cite[e.g.,][]{misra2024language}. Particularly relevant to our  question are studies that used Reinforcement Learning to investigate LMs' interactive learning strategies \cite{ma_babysit_2024, martins-etal-2025-upon, zhao-etal-2023-babystories, padovani2025dialogue, xu2025wanting, nikolaus_modeling_2021}.

Recent work has applied this approach to \textit{natural} child--caregiver interactions, using Reward Models trained from caregivers' actual feedback data \cite{nikolaus_fourtassi_2026}. Language models are then fine-tuned using techniques inspired by Reinforcement Learning from Human Feedback \cite[RLHF,][]{ouyang_training_2022}. This work showed that caregiver-like clarification requests on model generation can signal negative evidence. Learning from this signal improved grammaticality.


\subsection*{The current study} 
Existing RLHF-based studies of naturally occurring caregiver feedback have focused primarily on clarification requests, leaving open whether other forms of social feedback have similar, weaker, or even negative effects on grammatical learning. The present study addresses this gap by comparing several theoretically motivated feedback types within the same controlled learning framework.




\section{Methods}  
\label{sec:methods}

Following the broad methodology outlined in \citet{nikolaus_fourtassi_2026}, we use a two-stage learning setup designed to separate learning from linguistic exposure from learning driven by caregiver-feedback rewards.
In the first step, a small causal language model is pretrained on child-directed language (excluding children's production), providing an \emph{input-only} baseline that quantifies learning gains from linguistic exposure. In a second step, the model is fine-tuned via RL, using a reward model trained on the target caregiver's feedback valence. 
This design allows us to isolate factors that typically co-vary in children's experience; providing an opportunity to test their specific effects on learning. 


\noindent \textbf{Pretraining baselines}  We use a customized version of GPT-2 (2 layers, hidden size 512, 8 attention heads), pretrained on CHILDES \citep{macwhinney2000childes}, using three pretraining scales (0.1M, 1M, and 10M words, see Appendix 4).

\noindent \textbf{Empirical Reward Models} For each feedback type, we trained a reward model to capture naturally occurring patterns of caregiver feedback in child--caregiver interactions (using CHILDES dataset). Each reward model was trained to map children's utterances to a scalar, representing the valence of the target feedback (see Appendix 4). 
The valence values themselves were annotated prior to reward-model training using a range of automatic annotation tools (see below, and Appendix 3).

\noindent \textbf{Topline Reward Model} We use an idealized reward that directly targets grammatical errors. This topline is not intended to produce empirically realistic results, but rather to serve as an upper bound on what grammatical learning is possible \textit{in principle} (see details in Appendix 4).

\noindent \textbf{RL fine-tuning} For each feedback/reward, the pretrained models were fine-tuned using the corresponding reward models, using Proximal Policy Optimization \citep[PPO;][]{schulman_proximal_2017}. We included several safeguards to mitigate potential language drift (see Appendix 5).

\noindent \textbf{Training seeds and variability} To account for stochastic variability, we used, for each training scale, a fully crossed $3 \times 3 \times 3$ design over pretraining seeds, fine-tuning seeds, and generation seeds.

\begin{figure*}[t]
    \centering
    \includegraphics[width=\textwidth]{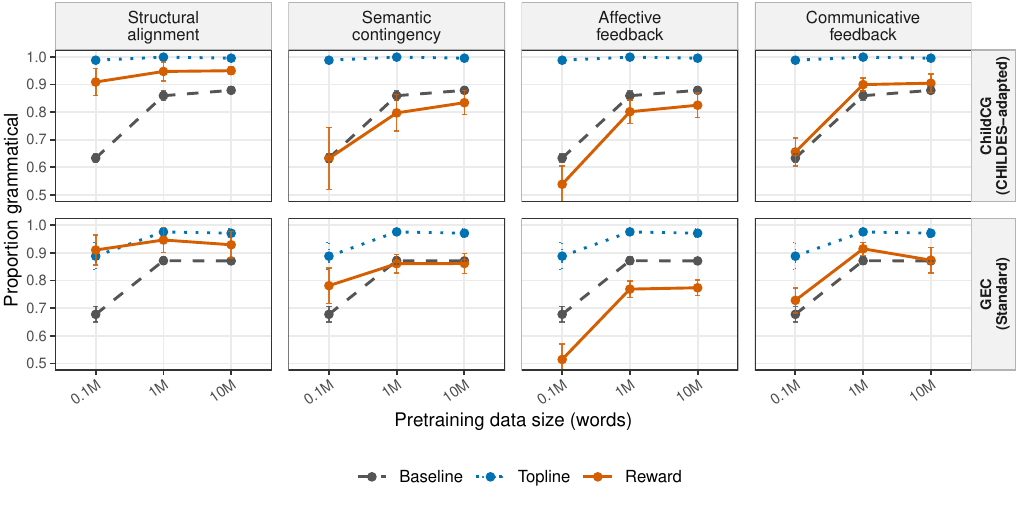}
    \caption{
    Generated-utterance grammaticality across pretraining scales for each reward.
    The top row shows results from the CHILDES-adapted ChildCG classifier, while the bottom row shows results using a standard GEC-based metric.
    Within each panel, baseline and topline trajectories are repeated to make the reward trajectory directly comparable.
    Means are computed over training-seed units after averaging over generation seeds. Error bars show $\pm 1$ standard deviation across training seeds.
    }
    \label{fig:generation_reward_scale}
\end{figure*}

\noindent \textbf{Annotation of the reward signals}  We considered the four classes of caregiver feedback mentioned in the introduction: communicative feedback, semantic contingency, structural alignment and affective feedback. For each type of feedback, we annotated the CHILDES child-caregiver conversations for the corresponding valence (See Appendix 3).






\subsection{Evaluation}
\label{sec:eval}

We use two complementary evaluations that capture different aspects of grammatical behavior. 

\noindent  \textbf{Minimal-pair benchmarks}: we use BLiMP \cite{warstadt2020blimp} and its CHILDES-adapted version Zorro \cite{huebner_babyberta_2021}. They test whether a model assigns higher probability to a grammatical sentence than to a minimally differing ungrammatical one. 

\noindent \textbf{Grammaticality of generated utterances} Free generations were scored using Child Conversational Grammaticality  \cite[ChildCG, ][]{nikolaus_automatic_2024}, a model adapted to CHILDES and its conversational context. In addition, we used a measure based on an off-the-shelf Grammatical Error Correction model \cite[e.g.,][]{rothe_simple_2021}. For each run and generation seed, we sampled up to 10,000 utterances using the Beginning-Of-Sentence (BOS) token as a prompt.


\section{Results}
\label{sec:results}

First, we compared the baseline models, trained only on linguistic input, with the same models further fine-tuned using the grammar topline reward model (Figure \ref{fig:topline} in Appendix). The topline improves both minimal-pair benchmark performance and the grammaticality of generation, showing the current framework can, in principle, lead to \textit{grammatical} learning above and beyond pretraining. 

Moving to the empirical reward models, we found no consistent improvement on the minimal-pair benchmarks (Figure~\ref{fig:benchmark_reward_scale} in Appendix). Across Zorro and BLiMP, none of the reward types reliably improved over the baseline. This echoes a broader pattern in recent studies showing
that interactive strategies  struggle to improve performance on these benchmarks (see also Discussion). In contrast, the generation results showed measurable effects on grammaticality (Figure \ref{fig:generation_reward_scale}). Across both ChildCG and GEC, the overall trends were largely consistent. Structural alignment produced the strongest and most consistent improvement across scales and metrics. Communicative feedback showed an overall positive effect, especially under ChildCG, but the gains were smaller. In contrast, semantic contingency and affective feedback generally reduced grammaticality below the baseline. 

Regarding the effect of pretraining size, the low-resource condition at 0.1M was visibly noisier, with larger variability across runs, excessively high non-word rates (Figure ~\ref{fig:covariates_reward_scale} in Appendix, bottom row), and overall low grammaticality scores. The effects became much more consistent from 1M words onward,
suggesting that a minimum amount of pretraining data is needed before reward-specific effects can emerge robustly.

We tested these qualitative patterns statistically using a mixed-effects logistic model. Predictors included reward type, pretraining scale, and their interaction. We additionally added the following predictors as controls: \uline{a) utterance length}, since longer utterances may naturally provide more opportunities for errors, and \uline{b) rate of non-words generated}, since out-of-dictionary forms may be penalized by grammaticality classifiers for reasons that are not strictly grammatical.  Finally, we included a random intercept for generation run to account for the fact that sampled utterances from the same seeds are not independent.

\begin{table}[t]
\centering
\small
\begin{tabular}{lll}
\toprule
Reward family & 1M & 10M \\
\midrule
Structural alignment      & $+.077^{***}$ & $+.060^{***}$ \\
Semantic contingency      & $-.092^{***}$ & $-.061^{***}$ \\
Affective feedback        & $-.038^{*}$   & $-.031^{*}$   \\
Communicative feedback    & $+.033^{*}$   & $+.024^{\dagger}$ \\
\quad Acknowledgment     & $+.017$       & $+.017$       \\
\quad Clarification request  & $+.050^{***}$ & $+.031^{*}$   \\
\bottomrule
\end{tabular}
\caption{
Model-estimated change in grammaticality probability relative to baseline, using the ChildCG metric at the 1M and 10M pretraining scales.
$\dagger p<.10$, $^{*}p<.05$, $^{**}p<.01$, $^{***}p<.001$.
}
\label{tab:grammar_reward_effects}
\end{table}

Table~\ref{tab:grammar_reward_effects} reports model-estimated changes in grammaticality probability relative to the baseline, using the more robust pretraining scales (1M and 10M). Mirroring the pattern in Figure~\ref{fig:generation_reward_scale}, structural alignment showed the clearest improvement. In contrast, semantic contingency and affective feedback both reduced grammaticality. Communicative feedback improved grammaticality at 1M and showed a weaker effect at 10M. When communicative feedback was split into its two components, only clarification requests showed a significant effect, whereas the acknowledgment-based reward model did not.


\section{Discussion}

We investigated which forms of caregiver feedback support grammaticality within an RLHF-inspired framework. We isolated and compared major feedback types, providing insight into their specific effects.

Our main finding is that structural alignment systematically improves grammaticality. Previous corpus work found that caregivers' alignment patterns predicted children's later language outcomes, with structural alignment specifically predicting increase in a measure of grammatical knowledge \cite{fusaroli_caregiver_2023}.
Complementing this line of work, the current result demonstrates, at scale, a plausible \textit{mechanistic} account: Caregivers' structural alignment can support grammar learning \textit{because} this reinforces the child's use of structures that are grammatical. This need not involve deliberate teaching: As competent speakers, caregivers are naturally more likely to re-use parts of children's utterances when they are well-formed than when they are ill-formed.

Regarding communicative feedback, it also improved grammaticality, though to a lesser extent. The improvement is consistent with prior work, especially on clarification requests \cite{nikolaus_fourtassi_2026, clark_conversational_2020}. 

For affective feedback and semantic contingency, neither improved grammaticality; instead, both reduced it.
Echoing our finding, previous research on parenting behavior did not find a specific relation between affective behavior and language development \cite{rivero2023relations}. The authors suggested that, on its own, this type of feedback may be less directly related to language learning than to socio-cognitive development.  Our supplementary analyses nevertheless suggest that affective feedback may still support some aspects of language development, though not necessarily in grammar. For instance, results in Figure \ref{fig:covariates_reward_scale} and model-generated samples (Table \ref{tab:sampled_childcg_examples}) show that affective feedback increased utterance length relative to the baseline. This suggests that it may encourage longer vocalizations, which---while increasing opportunities for grammatical errors---can potentially support early phonological development \cite{goldstein2003social, nikolaus_communicative_2022, warlaumont_social_2014}.

For semantic contingency, supplementary analyses (Figure \ref{fig:covariates_reward_scale}) show increased content-word use  relative to baseline and greater lexical diversity than the other feedback types. These effects are consistent with previous studies linking caregivers’ semantic and lexical contingency to children’s subsequent lexical growth and diversity \cite[e.g.,][]{fusaroli_caregiver_2023, che2018assessing}.  

As for why semantic contingency reduced grammaticality, one possible explanation is that our measure captured two competing mechanisms. Our original hypothesis was that high semantic contingency signals communicative success, which should be more likely when children produce well-formed utterances that caregivers can readily understand \cite{nikolaus_communicative_2023}. However, high semantic contingency can also arise when caregivers reformulate children’s errors \cite{chouinard_adult_2003}. Ungrammatical utterances may therefore elicit semantically contingent responses, as caregivers reformulate the incorrect form while preserving the intended meaning. A reward model favoring such responses could inadvertently reinforce ungrammatical productions. Future work should disentangle these mechanisms.

Thus, different types of feedback push generation in different directions: some improve grammaticality, while others encourage other properties of language use. 
Future work should investigate how these feedback types may support \textit{complementary} aspects of language learning beyond grammar.

\subsection*{Minimal-Pair vs. Generation-Based Evaluation}

The topline reward improved on baseline in grammatical benchmarks, indicating the viability of this learning framework (Figure \ref{fig:topline}). However, the effects of the empirical rewards were observed in model generation (Figure \ref{fig:generation_reward_scale}), but not in minimal-pair benchmarks (Figure \ref{fig:benchmark_reward_scale}). This echoes a broader pattern in recent studies showing that interactive strategies generally struggle to improve performance on minimal-pair benchmarks of grammar such as BLiMP and Zorro \cite{zhao-etal-2023-babystories, padovani2025dialogue, nikolaus_fourtassi_2026, stopler2025towards} and work showing that child linguistic environment has more apparent effects in production-based evaluation than on minimal-pair benchmarks \cite{bunzeck2026child}.  

This raises an important question: To what extent can changes in grammatical production be taken as evidence of changes in underlying grammatical knowledge? This echoes a long-standing debate in child language research \cite[e.g.,][]{tomasello2000young, fisher2002role}

In the present study, we specifically found that learning from caregiver-like reward changes the model’s generative behavior. 
The generation-based results are informative for theories of language development. Unlike  corpus-level correlations, the rewards here were used as interventions within a learning system, allowing us to test whether optimizing for these signals \textit{causally} shifts the model’s production policy in grammar-relevant directions.

That said, a model becoming more grammatical in free generation does not necessarily mean that it has learned new grammatical regularities---or has unlearned ungrammatical patterns---in a way that generalizes beyond the specific utterance types rewarded during fine-tuning.
Minimal-pair benchmarks---though not perfect \cite{martinez2023evaluating, weissweiler2025linguistic}---are useful precisely because they provide out-of-distribution tests for general learning. Our null results on BLiMP/Zorro (Figure \ref{fig:benchmark_reward_scale}) therefore matter. They suggest that the empirical rewards did not produce grammatical generalization---at least, not of the kind captured by these benchmarks.

Although this interpretation is plausible, it would be premature to draw it here without further research. Existing benchmarks do not exhaust the space of possible grammatical generalizations. Even child-adapted benchmarks such as Zorro primarily target classic grammatical phenomena, which overlap only partially with errors common in child \textit{production} \cite{saxton_prompt_2005, hiller_data-driven_2016, nikolaus_automatic_2024}. This is relevant because in RL, errors in productions are the ones that receive direct pressure from the reward.

Testing this learning claim, thus, remains a direction for future work. It requires conducting an exploration of the grammatical errors found in both child and model productions, and using these bottom-up insights to build targeted tests around the relevant phenomena.  


\subsection*{Limitations}

A limitation of this work is that we captured only the verbal component, which could have under-estimated the learning challenge associated with credit assignment \cite{pinker_formal_1979, marcus_negative_1993}. In reality, children must interpret feedback in real time within a multimodal context \cite{, behne2012twelve, adamson2004development, fourtassi2020optimal}. This makes credit assignment especially challenging, because the learner must disambiguate among several possible sources of error beyond the combinatorial structure of the child's utterance. Valenced feedback may reflect not only how symbols are combined, but also a mispronounced word or an inappropriate lexical choice given the visual context.

At the same time, restricting our analysis to verbally expressed caregiver feedback may have underestimated the richness of the feedback available to children. Caregivers can signal whether and how they have understood a child through multiple channels, including linguistic responses, prosody, gaze, gesture, and facial expression.

A second limitation of this work is that we used a single GPT-2-style architecture. Although we used a fully crossed design over pretraining, RLHF fine-tuning, and generation seeds to account for multiple sources of variability, we did not establish generality across model architectures, mainly because duplicating this large set of experiments across several architectures would require substantial computational resources.

This limitation, howevever, should be interpreted in light of our objective. We do not propose a general-purpose method for improving language models; rather, we use a controlled child-like neural learner as an experimental testbed to examine whether naturally occurring caregiver feedback contains information that can support grammatical development. Nevertheless, replication across architectures would strengthen the generality of this conclusion.

Another potential concern is that reward fine-tuning could narrow the model’s production distribution, reducing diversity and potentially harming generalization. 
Our implementation incorporates several safeguards against an excessive loss of diversity, including entropy and language-modeling regularization, adaptive KL control to limit divergence from the pretrained distribution, and randomly sampled short prompts that encourage exploration across a broad range of constructions during the fine-tuning stage (see Appendix~5).

However, some narrowing of the production distribution is both expected and consistent with learning if the model shifts away from ill-formed constructions, effectively ``unlearning'' ungrammatical options---the primary hypothesized role of negative evidence in grammar \citep{saxton_contrast_1997}.

For instance, in our analyses, the grammar topline was relatively the least diverse condition (Figure~\ref{fig:covariates_reward_scale}, Lexical Entropy, Appendix), yet it produced the clearest gains on the minimal-pair benchmarks (Figure~\ref{fig:topline}, Appendix). Thus, reduced diversity need not preclude grammatical generalization and may sometimes accompany it. 

Finally, beyond final performance, interactive feedback may also shape learning trajectories. For example, \citet{ma_babysit_2024} show that interaction can reduce the amount of data or experience required to reach comparable performance in word learning.   \citet{nikolaus_modeling_2021} found that feedback-based models align with the developmental order of several aspects of visually grounded grammar learning.  A direction for future work is to investigate these dynamics within a framework such as the one outlined here, in which realistic caregiver feedback serves as a reward signal.

\section*{Ethical considerations}

This work uses existing child-caregiver corpora and does not involve new data collection. Because the data involve children, we treat the corpora as sensitive and report only aggregate model-level results, without attempting to identify individual children, caregivers, or families. The proposed modeling framework is intended as a tool for studying hypotheses about language development, not for evaluating individual children or caregivers.

\section*{Acknowledgments}

This work was supported by the ANR MACOMIC project (ANR-21-CE28-0005-01) and was carried out within the Institut Convergence ILCB (ANR-16-CONV-0002). This work was performed using HPC and storage resources from GENCI--IDRIS (Grant 2025-AD011013886R3) on the Jean Zay supercomputer.

This work has received funding from the European Union's Horizon 2020 research and innovation programme under the Marie Skłodowska-Curie grant agreement No 945304 – Cofund AI4theSciences hosted by PSL University.

\bibliography{custom}

\appendix

\clearpage
\twocolumn[{
\section*{Appendix 1: Supplementary Analyses}
\label{app:analyses}

\begin{center}
\includegraphics[width=0.7\textwidth]{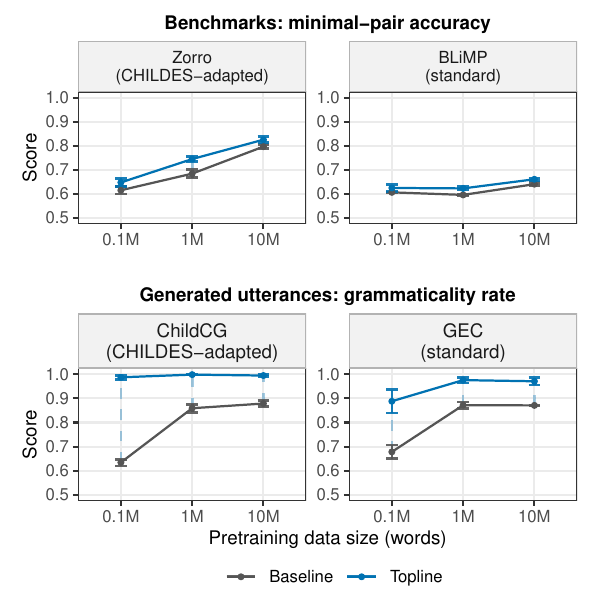}

\captionof{figure}{
Baseline and Topline performance across pretraining scales. The top row reports minimal-pair benchmark accuracy for Zorro and BLiMP. The bottom row reports grammaticality of generated utterances, using the ChildCG or GEC-based metric. Points show means across training seeds (after averaging over generation seeds). Error bars show ±1 standard deviation across these independent units.
}
\label{fig:topline}
\end{center}
}]

\begin{figure*}[t]
    \centering
    \includegraphics[width=\textwidth]{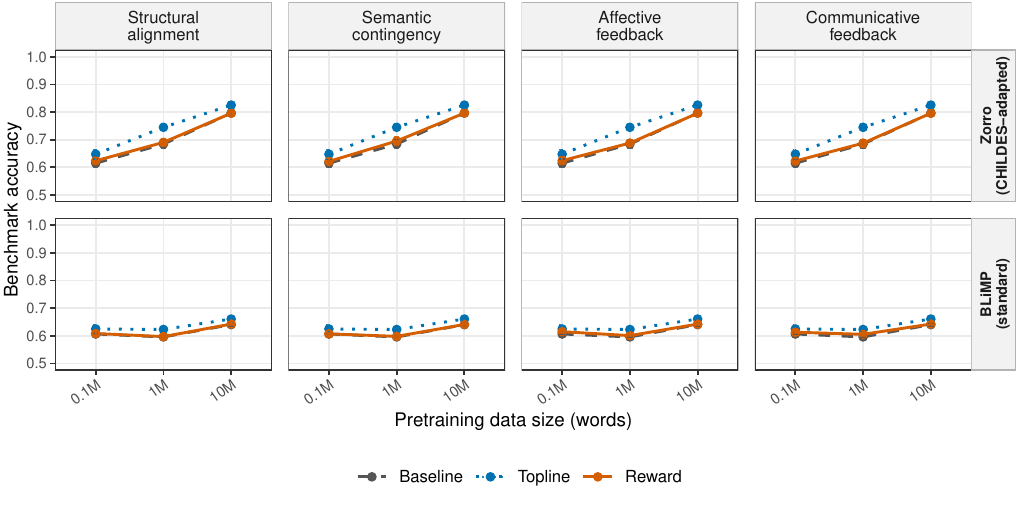}
    \caption{
    Benchmark performance across pretraining scales for each reward family.
    Rows show Zorro and BLiMP minimal-pair accuracy; columns show the four empirical reward families.
    Within each panel, the baseline and topline trajectories are repeated to make the reward trajectory directly comparable at each scale.
    Means are computed over training seeds, after averaging over generation seeds. Error bars show $\pm 1$ standard deviation across independent units.
    }
    \label{fig:benchmark_reward_scale}
\end{figure*}

\begin{figure*}[t]
    \centering
    \includegraphics[width=\textwidth]{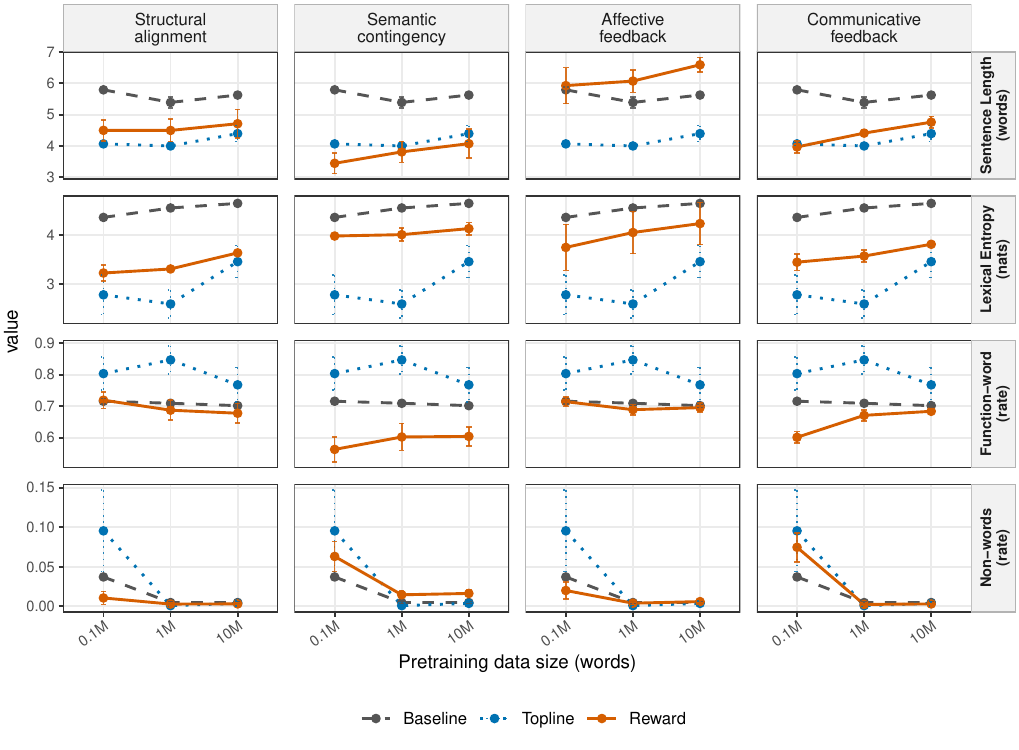}
    \caption{
    Properties of generated utterances: Sentence length, pooled lexical entropy, function-word density, and rate of non-words.
    Points show means for each reward across training seeds, with error bars indicating $\pm 1$ standard deviation across these seeds.
    Within each panel, baseline and topline trajectories are repeated to make the reward trajectory directly comparable at each scale.
    }
    \label{fig:covariates_reward_scale}
\end{figure*}

\clearpage

\begin{table*}[t]
\section*{Appendix 2: Examples of Model-Generated Utterances}
\vspace{1em}
\centering
\scriptsize
\setlength{\tabcolsep}{3.5pt}
\renewcommand{\arraystretch}{1.18}

\begin{tabular}{
p{0.12\textwidth}
p{0.12\textwidth}
p{0.30\textwidth}
>{\columncolor{gray!15}}p{0.38\textwidth}
}
\toprule
Reward type &  & ChildCG = 1 (grammatical) & ChildCG = 0 (ungrammatical) \\
\midrule

\textbf{Baseline} &  &

\textit{you want to look at some more pictures?}

\textit{oh, is that what it looks like?}

\textit{it is a big boy now isn't it?}

\textit{have you not any other little boys?}
&
\textit{this piece of cake?}

\textit{because you can see mummy's poorly sick.}

\textit{the dog is saying.}

\textit{he is the name of the little girl.}
\\
\midrule

\textbf{Topline} &  &
\textit{what are you doing?}

\textit{there is a horse.}

\textit{that is a girl.}

\textit{are you going back?}

\textit{that is a chair.}

\textit{what is that one?}

\textit{i will put it there.}

\textit{what did we do with the barn?}
&
\textit{why is that side?}
\\
\midrule

\begin{tabular}[t]{@{}l@{}}\textbf{Structural}\\\textbf{Alignment}\end{tabular} & &
\textit{so that is a duck.}

\textit{he is a turtle.}

\textit{you are a good boy.}

\textit{i am a monkey.}
&
\textit{how many trucks in the sky.}

\textit{come down the street.}

\textit{we are a good boy.}

\textit{can't find it.}
\\
\midrule

\begin{tabular}[t]{@{}l@{}}\textbf{Semantic}\\\textbf{Contingency}\end{tabular} & &
\textit{yes that is a doggie.}

\textit{well it is a boat.}

\textit{there is an elephant.}

\textit{yes a banana puppy.}
&
\textit{where is baby sheep?}

\textit{what is big truck?}

\textit{he a lion.}

\textit{a big flower red.}
\\
\midrule

\begin{tabular}[t]{@{}l@{}}\textbf{Affective}\\\textbf{feedback}\end{tabular}
&  &
\textit{and then i will put it in it.}

\textit{i will put it over here if i put it in there.}

\textit{we have to wash her hair first okay?}

\textit{we will shall we put it all together?}
&
\textit{if i do it and then we can do it and put it in?}

\textit{now i will have to clean this with your room won't you?}

\textit{want mama do it?}

\textit{because your teeth i am watching this picture aren't they?}
\\
\midrule

\begin{tabular}[t]{@{}l@{}}\textbf{Communicative}\\\textbf{feedback}\end{tabular}
& Acknowledgment &
\textit{they are called flowers.}

\textit{that is a gold straw.}

\textit{and that is the baby.}

\textit{you have a pink one on there.}
&
\textit{have to use your spoon and pretend.}

\textit{you will use them ones with this bits.}

\textit{all day and then the is warm.}

\textit{but two grey stars.}
\\
\cmidrule(lr){2-4}

& \begin{tabular}[t]{@{}l@{}}Clarification\\request\end{tabular} &
\textit{i don't know.}

\textit{can you see it?}

\textit{can't you?}

\textit{mhm, it is.}
&
\textit{look at that isn't it?}

\textit{look it, there is it?}

\textit{see you shouldn't?}

\textit{there doesn't.}
\\

\bottomrule
\end{tabular}

\caption{
Examples of utterances generated by RL-fine-tuned models at the 10M-word scale (generation was qualitatively similar at the 1M-word scale).
Examples are organized by reward types. Unshaded cells show utterances that were classified as grammatical by ChildCG, while shaded cells show utterances classified as ungrammatical.
}
\label{tab:sampled_childcg_examples}
\end{table*}

\clearpage
\section*{Appendix 3: Automatic labeling of rewards}

We automatically annotated caregiver feedback signals in child-caregiver conversations and used these annotations to supervise reward model training (see Appendix 4). 

\subsection*{Communicative feedback}

\noindent \textbf{Clarification requests} 
Clarification requests were identified using a DeBERTa-v3-xsmall classifier fine-tuned on manually annotated caregiver responses in CHILDES, following prior work \citep{nikolaus_fourtassi_2026}.

\noindent \textbf{Acknowledgments} To identify caregiver acknowledgments, we relied on the feature-based algorithm described in \citet{nikolaus_communicative_2023a}, which was specifically designed for CHILDES data and combined common keywords (e.g., ``okay'', ``alright'', and ``yeah'') as well as repetition ratios capturing repetition-based acknowledgments (e.g., ``It isn’t very nice, is it?'' – ``It isn't.''). 

\subsection*{Structural Alignment}


Following previous research \cite{dale_unraveling_2006, fernandez_quantifying_2014}, a child structure was defined as a POS bigram. We used spaCy (\texttt{en\_core\_web\_sm})\footnote{\url{https://huggingface.co/spacy/en_core_web_sm}} to tag each child utterance and caregiver response for part-of-speech (POS), excluding punctuation and empty tokens. Caregiver alignment was computed as a normalized score of bigram reuse, corresponding to the size of the intersection between the child and caregiver POS-bigram sets, divided by the size of the larger set.

\subsection*{Semantic Contingency}

Following previous research \cite[e.g.,][]{fusaroli_caregiver_2023, foushee2022getting}, semantic contingency was computed using sentence embedding similarity. Here, we used \texttt{all-MiniLM-L6-v2}\footnote{\url{https://huggingface.co/sentence-transformers/all-MiniLM-L6-v2}} to encode the child utterance and caregiver response. We computed the cosine similarity between the two embeddings, and clipped the resulting score to the $[0,1]$ range.

\subsection*{Affective feedback}

We annotated affective properties of caregiver responses using \texttt{roberta-base-go\_emotions}.\footnote{\url{https://huggingface.co/SamLowe/roberta-base-go_emotions}} The model was applied to adult responses only. For each caregiver utterance, the classifier returned a probability distribution over several labels. We used the probabilities for emotions related to supportiveness, approval, and warmth.



\section*{Appendix 4: Baseline and Reward model training}

\subsection*{Baseline models}
For the input-based baselines, we trained a small GPT-2-style causal language model on caregiver utterances only. The model used 2 hidden layers, 8 attention heads, and a hidden size of 512. We trained a word-level tokenizer on the training data, retaining words that appeared at least twice and capping the vocabulary at 5,000 types. Models were optimized with the standard GPT-2 causal language-modeling objective using AdamW and a batch size of 256. For each data scale, we held out 10\% of the data for validation and used validation loss for early stopping and checkpoint selection. To assess the effect of input size, we trained models on three amounts of caregiver language: 0.1M, 1M, and 10M words. Each condition was run with three different random seeds.

\subsection*{Reward models}

For each of the four caregiver feedback dimensions above, we trained a reward model that maps a child utterance to the corresponding valence of the caregiver response, as annotated using the procedures described in Appendix 3.


All reward models share the same pretrained model:
\texttt{microsoft/deberta-v3-xsmall} \citep{he_debertav3_2022}, which we fine-tuned using a linear regression layer and a mean squared error (MSE) loss to predict the target reward for a given utterance.

All reward models were trained on the CHILDES child--caregiver conversation pairs (one row per child utterance with the immediately following caregiver response), the same pre-processed  pipeline above. 
We split the data into 90\% training and 10\% held-out evaluation with a fixed random seed.  
We used the AdamW optimizer with an initial learning rate of $1.41 \times 10^{-5}$, a batch size of 128, and early stopping based on MSE on the held-out validation set.
The best checkpoint is selected on held-out MSE and used as the frozen reward model during PPO fine-tuning.

\subsection*{Topline Reward Model}

The topline reward model was obtained using the same procedure, except that instead of learning the mapping between child utterances and the feedback valence, it was trained on the controlled minimal-pair datasets used in Zorro and BLiMP. More specifically, the model was trained to classify sentences from these datasets according to their labels (i.e., grammatical or ungrammatical).

\section*{Appendix 5: RL Fine-tuning Details}

After pretraining on child-directed CHILDES utterances, we fine-tune each language model with Proximal Policy Optimization \citep[PPO;][]{schulman_proximal_2017}, implemented using Hugging Face’s TRL library.
For each fine-tuning step, we (a) sample utterances from the current language model, (b) compute the corresponding rewards using the reward model, and (c) update the model weights using PPO.

To obtain a diverse set of generated utterances, we randomly prompted the model with short utterance prefixes (the first one or two tokens) sampled from the pretraining data, namely the caregiver utterances used to train the input-based baseline model.
We additionally included an entropy regularization term (0.001) in the loss. 

The models were fine-tuned for a maximum of 6,000 steps, and the best checkpoint was selected based on mean reward. To discourage excessively short or long utterances, we applied rejection sampling by assigning a reward of $-1$ to generated utterances that were shorter than three tokens or failed to produce an end-of-sequence token within 20 tokens.

To mitigate language drift, a known issue in Reinforcement Learning (RL) fine-tuning, we also added a small language-modeling regularization term (weight = 0.001). We used the TRL default optimizer and learning rate. Other hyperparameters are shown in Table \ref{tab:ppo-hparams}. All other PPO hyperparameters were not changed from the default values as implemented in the Huggingface TRL library.


Each PPO run uses one NVIDIA H100 GPU (96\,GB) with bf16 mixed precision and completes in approximately \textsc{6}\,GPU-hours.

\begin{table}[t]
\centering
\small
\begin{tabular}{ll}
\toprule
\textbf{Hyper-parameter} & \textbf{Value} \\
\midrule
Max PPO total steps                 & 6000 \\
Batch size                       & 1024 \\
Mini-batch size                  & 512 \\
KL controller                    & adaptive, target $\text{KL}^\star=6$ \\
Entropy coef.\ $c_e$             & $1\times10^{-3}$ \\
LM mixing coef.\ $\lambda_{\text{LM}}$ & $1\times10^{-3}$ \\
Score clipping                   & disabled \\
Sampling                         & $T{=}1.0$, top-$p{=}1.0$, top-$k{=}0$ \\
Precision                        & bf16 mixed \\
Eval frequency                   & every 100 steps \\
Log frequency                    & every 10 steps \\
Early stopping                   & no reward gain for 10 log windows \\
\bottomrule
\end{tabular}
\caption{PPO fine-tuning hyper-parameters, shared across the four caregiver-feedback rewards, the grammar topline, and all pretraining scales.}
\label{tab:ppo-hparams}
\end{table}



\end{document}